# Visual Concept Detection and Real Time Object Detection

Ran Tao
LIACS Media Lab

## ABSTRACT
Bag-of-words model is implemented and tried on 10-class visual concept detection problem. The experimental results show that "DURF+ERT+SVM" outperforms "SIFT+ERT+SVM" both in detection performance and computation efficiency. Besides, combining DURF and SIFT results in even better detection performance. Real-time object detection using SIFT and RANSAC is also tried on simple objects, e.g. drink can, and good result is achieved.

## 1. INTRODUCTION
Recent years have witnessed an explosive growth of available digital images. For instance, Flickr, the photo sharing website, hosts billions of images and has millions of images uploaded every month. It becomes impossible to annotate such huge collections of images manually. It urges for an efficient and reliable way of automatic image annotation. Visual concept detection and matching [1, 5, 12, 13] is about the prediction of the presence/absence of a visual concept in images and can annotate images automatically.

Visual concept detection is currently a hot research topic in image retrieval and computer vision communities. Although great progress has been achieved, it is still very imprecise, comparing with manual annotations. The main difficulty lies in the so-called "bridging the semantic gap" [1]. Briefly, two images of the same concept can be very different in low level visual features. A concept detector has to learn such differences.

The rest of this report is structured as follows. Section 2 is some related work. Section 3 describes the approach. Experimental results are presented in Section 4. Section 5 concludes.

## 2. RELATED WORK
The state-of-the-art method in visual concept detection is based on the bag-of-words model. This model is perceived as a pipeline of four components: Sampling Strategy, Visual Feature Extraction, Word Assignment and Concept Learning.

Lowe [2] proposed a salient point detector based on scale space theory and the difference of Gaussian. Bay *et al.* [3] introduced a Fast-Hessian interest point detector, which is more

Nowak *et al.* [4] showed that sampling on a regular dense grid outperforms complex salient point detectors[7] simply because dense sampling produces much more image patches than interest point operators. computationally efficient by using integral images and box filters.

SIFT descriptor [2] and SURF descriptor [3] are two commonly used image patch descriptors. Both of them are rotation-invariant and illumination independent.

Word assignment is the process of assigning descriptors to a visual vocabulary and each descriptor is mapped to a visual word. A common method of creating large visual vocabularies is unsupervised k-means clustering which gives good performance (e.g. in [5]). Moosmann *et al.* [6] proposed a tree-based method for codebook generation and word assignment, named "Extremely Randomized Clustering Forests", which brings a significant computational advantage and gives comparable performance in image classification tasks.

Support vector machine (SVM) is successfully used in the bag-of-words model (e.g. [4] [5] ).

## 3. THE APPROACH
### 3.1 Visual Concept Detection
Specifically, the visual concept detection problem solved here is defined as: For each class, predicting the presence of an example of that class in test images. The bag-of-words model is applied and the specific approaches for each component of the pipeline are described in the following subsections.

#### 3.1.1 Dense Sampling
Dense sampling is employed, which samples points on a regular dense grid, i.e. points are sampled using a fixed pixel interval. Different from the multi-scale sampling in [4], points are sampled only on one scale in our method. In the experiments, an interval distance of *6s* pixels is used, where *s* refers to the sampled scale.

#### 3.1.2 Descriptor
For each sampled point, a SURF-like descriptor is extracted using its neighboring pixels. The first step is to construct a square window of size *24s* around the sampled point, again where *s* refers to the sampled scale. Then this descriptor window is divided regularly into 4*4 subregions, each of which is described by the summation of pixel-wise Haar wavelets of size *2s*. Thus, each subregion is described by a 4-dimensional vector ($\sum dx, \sum dy, \sum |dx|, \sum |dy|$), where *dx* and *dy* refer to the x and y wavelets responses respectively. And the feature descriptor of this sampled point is the concatenation of these 16 4-dimensional vectors, i.e. of dimension 64.

There exist two differences between the SURF-like descriptor used in our method, as described above, and the traditional SURF descriptor [3]. Unlike [3], the step of orientation assignment is not included in our method, i.e., the property of rotation-invariant does not hold. Though rotation invariance is a necessary property for object matching, it is not that important for large scale visual concept detection task and the remove of orientation assignment brings computational advantage. The second difference is that in our method we remove the Gaussian weighting centered at the sampled point, while in [2] [3] Gaussian weighting is applied to increase the robustness toward interest point localization errors. The reason for removing this is simply that in our method, dense sampling is used, instead of interest point detection.

The SURF-like descriptor, together with the dense sampling, is named "DURF"[1].

### 3.1.3 Codebook Generation and Word Assignment

"Extremely Randomized Trees" (ERT) [6] [8] is applied for codebook generation and word assignment. The resulting codebook consists of several binary trees and the leaves of these trees are indexed, serving as visual words.

### 3.1.4 Visual Concept Learning

SVM is used for visual concept learning and histogram intersection kernel (HIK) is applied. Histogram intersection was first introduced in [9] and HIK is defined as

$$K(x,z) = \sum_{i=1}^{k} \min(x_i, z_i) \quad (1)$$

HIK-SVM classifier is trained for each visual concept. The output of the trained classifier on the test image is the real-valued confidence of the concept's presence in the image.

### 3.1.5 Implementation

The bag-of-words pipeline has been implemented as a C++ library. The simplified class diagram is shown in Figure 1.

As shown in Figure 1, the components of Sampling Strategy and Visual Feature Extraction are represented by the abstract class *OpenVCDDescriptorExtractor*. Class *OpenVCDDurfDescriptor-Extractor* and class *OpenVCDSiftDescriptorExtractor* are inherited from the abstract class *OpenVCDDescriptorExtractor*. *OpenVCDDurfDescriptorExtractor* is for the dense sampling and SURF-like descriptor extraction described above, while *OpenVCDSiftDescriptorExtractor* is for the SIFT feature extraction in [2]. The C++ abstract class *OpenVCDDictionary* is for the component of Word Assignment. Class *OpenVCDERTreesDictionary*, inherited from abstract class *OpenVCDDictionary*, is for the "Extremely Randomized Trees" algorithm. The abstract class *OpenVCDDetector* is for the component of Concept Learning. Class *OpenVCDSVM* is the implementation of SVM. Such a design is flexible. Different methods of the bag-of-words pipeline components can be implemented and added into the library easily, as long as they inherit from the corresponding abstract class. And with these abstract classes, the interface can be unified, e.g. *OpenVCDDurfDescriptorExtractor* and *OpenVCDSiftDescriptor-Extractor* have the same interface function for point sampling and image patch descriptor extraction, "*void extract(const string imagePath, vector<OpenVCDKeyPoint> & keyPoints) const;*".

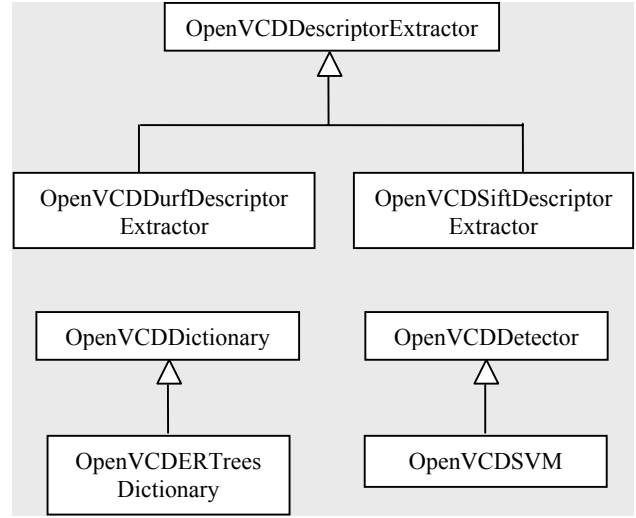

Figure 1. Simplified Class Diagram

## 3.2 Real Time Object Detection

The problem of real time object detection solved here is defined as learning an object and finding it in real time through a webcam input. The whole process can be divided into the following steps:

1. Get an image of the target object through the webcam; detect the salient points and extract the visual feature descriptors with the SIFT algorithm [2]; (object learning)

2. For the webcam input frame, detect the salient points and extract descriptors;

3. Find the corresponding points between the input frame and the target object using the approximate nearest-neighbor search algorithm proposed in [11];

4. With the corresponding point pairs, use the RANSAC approach [10] to compute the homography if possible, and then locate the object in the input video using the computed homography.

### 3.2.1 Implementation

A Windows application has been implemented. DirectShow SDK is used for video input device selection and real time video capture and rendering.

Screenshots of detecting a 7-up can are shown in Figure 2.

---

[1] The name of "DURF" is essentially a denser version of SIFT[2]. In Table 1 and Figure 3, DURF refers to "DURF+ERT+SVM", SIFT refers to "SIFT+ERT+SVM" and DURF&SIFT refers to "DURF&SIFT+ERT+SVM". But in other places within this article, DURF refers to dense sampling together with SURF-like descriptor and SIFT refers to the salient point detection and SIFT descriptor in [2].

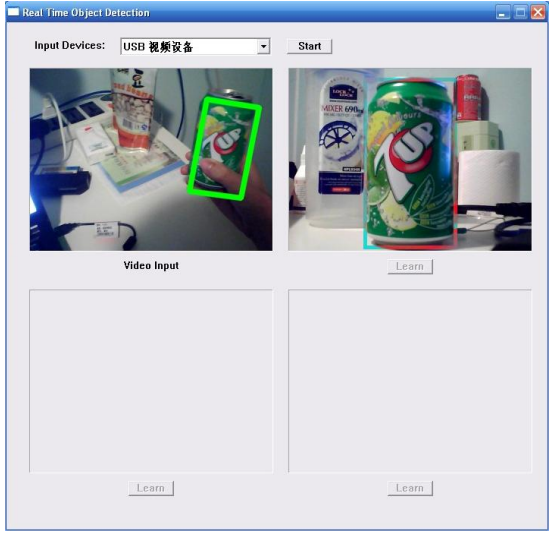

**Figure 2(a). The can is learnt once.**

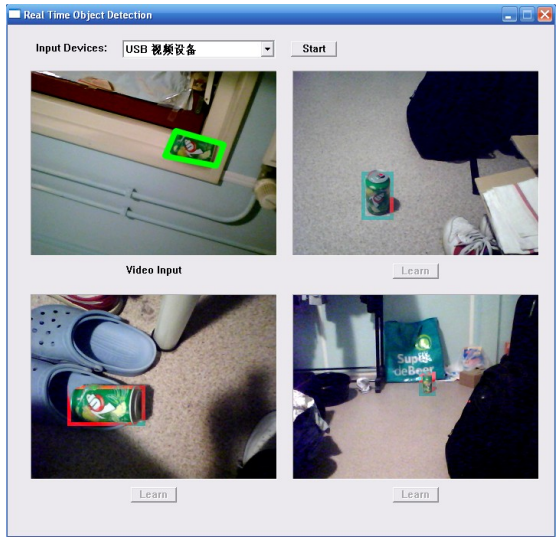

**Figure 2(b). The can is learnt three times.**

## 4. EXPERIMENTAL RESULTS

The "DURF+ERT+SVM" approach for visual concept detection is compared with the "SIFT+ERT+SVM" method in the aspects of both detection performance and computation efficiency. Another method, named as "DURF&SIFT+ERT+SVM", is also compared, which combines the visual word representations of the previous two approaches.

I **Dataset.** The used dataset is the Pascal VOC 2007 dataset, consisting of 9963 images divided into 3 predefined sets: "train set" (2501 images), "val set" (2510 images) and "test set" (4952 images). The "test set" of 4952 images is used to generate the codebook. The "train set" is used for concept learning and part of the "val set" is used for testing.

II **DURF.** The sampling scale of dense sampling strategy is set to be 2, and the interval distance is 6*2(=12) pixels.

III **Codebook.** In the "DURF+ERT+SVM" approach, for each of the 4952 images, from all the sampled points on this image, 100 points are randomly selected and the SURF-like descriptor is extracted for each point. The resulting 64-dimensional descriptors are used to generate the codebook, which consists of 4 binary trees with a maximal depth of 12. The size of the codebook is 7840.

In the "SIFT+ERT+SVM" method, for each of the 4952 images, 100 points are randomly selected from all the detected salient points and the SIFT descriptors [2] are extracted. The resulting 128-dimensional descriptors are used for codebook creation. The codebook is of size 7975, again consisting of 4 binary trees with a maximal depth of 12.

IV **Concept Learning.** For each of the 10 visual concepts, i.e. *aeroplane*, *bird*, *bottle*, *car*, *chair*, *diningtable*, *horse*, *person*, *sofa and tvmonitor*, a HIK-SVM classifier is learned with the "train set". For instance, for the concept *aeroplane*, 112 positive images and 2388 negative images from the "train set" are used to learn the classifier.

V **Testing.** For each of the 10 visual concepts, 100 positive images and 900 negative ones, randomly selected from the "val set", are used to test the learned classifier. The final output for each concept is a ranked list of the 1000 images, according to the confidence of the concept's presence in the image.

Average Precision (AP) is calculated to measure the detection performance for each concept. In the aspect of computation efficiency, SIFT [2] and DURF are compared by recording the time for extracting the descriptors of the same image set of size 1000. The final detection performance measured in AP is presented in Table 1.

**Table 1. Detection Performance of the Three Approaches**

|  | **DURF** | **SIFT** | **DURF&SIFT** |
|---|---|---|---|
| **Aeroplane** | 0.7193 | 0.5497 | 0.7197 |
| **Bird** | 0.2996 | 0.2657 | 0.4092 |
| **Bottle** | 0.2955 | 0.1978 | 0.2971 |
| **Car** | 0.5496 | 0.2873 | 0.5855 |
| **Chair** | 0.3674 | 0.2446 | 0.3983 |
| **Diningtable** | 0.3151 | 0.2213 | 0.3354 |
| **Horse** | 0.4584 | 0.3271 | 0.5069 |
| **Person** | 0.2922 | 0.2395 | 0.3104 |
| **Sofa** | 0.3901 | 0.1975 | 0.3891 |
| **Tvmonitor** | 0.4229 | 0.3382 | 0.4689 |
| **MAP[2]** | 0.4110 | 0.2869 | 0.4421 |

Obviously, the "DURF+ERT+SVM" approach outperforms "SIFT+ERT+SVM", even though DURF only samples on one scale and the SURF-like descriptor is not rotation-invariant. It is clear that the number of sampled image patches is a significant parameter governing the detection performance. The better result achieved by "DURF&SIFT+ERT+SVM" indicates that DURF and SIFT are complementary and even better result can be expected if other sampling strategies and feature extraction algorithms are combined.

---

[2] MAP: mean average precision over all classes.

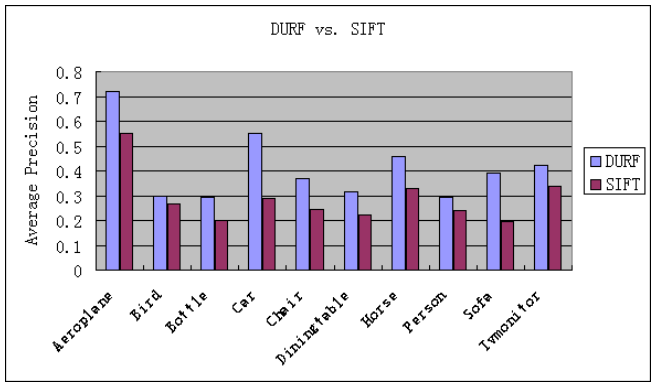

**Figure 3(a). Detection Performance of DURF and SIFT**

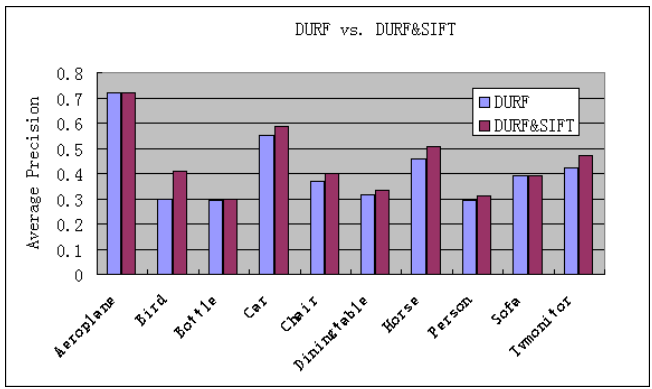

**Figure 3(b). Detection Performance of DURF and DURF&SIFT**

Besides better detection performance, DURF is also more computationally efficient. DURF is approximately 10 times faster than SIFT in extracting the descriptors of 1000 images.

## 5. CONCLUSIONS AND FUTURE WORK

For the 10-class visual concept detection problem, the approach of "DURF+ERT+SVM" achieves much better detection performance than "SIFT+ERT+SVM", which shows that the number of sampled image patches is a significant factor. And by using dense sampling and SURF-like descriptor, DURF is approximately 10 times faster than SIFT in feature extraction. Combining the DURF and SIFT gives the best result in the experiments, which indicates that even better result can be expected if other sampling strategies and feature extraction algorithms are combined.

The detection performance on the concepts that have great intra-class difference, such as *person* and *bird*, is not satisfactory. Adding spatial information into the bag-of-words model might increase the detection performance on these complex classes, which is the future work.

For real time object detection, due to the learning time constraint and the limitation of the applied algorithm, it only works well for simple objects, such as 7-up can and playing card. Further research work will be done to extend its detection power.

## 6. ACKNOWLEDGMENTS
I would like to thank Dr. Michael S. Lew for giving me many valuable advices and comments.

## 7. REFERENCES

[1] M. S. Lew, N. Sebe, C. Djeraba, and R. Jain, "Content-based multimedia information retrieval: state of the art and challenges," *ACM Transactions on Multimedia Computing, Communications, and Applications*, 2(1): 1-19, February, 2006.

[2] D. Lowe, "Distinctive image features from scale-invariant keypoints," *International Journal of Computer Vision*, vol. 60, pp. 91-110, 2004.

[3] H. Bay, A. Ess, T. Tuytelaars, and L. V. Gool, "Speeded-up robust features (surf)," *Computer Vision and Image Understanding*, vol. 110, pp. 346-359, 2008.

[4] E. Nowak, F. Jurie, and B. Triggs, "Sampling strategies for bag-of-features image classification," *Proc. Ninth European Conf. Computer Vision*, 2006.

[5] J. Zhang, M. Marszałek, S. Lazebnik, and C. Schmid, "Local features and kernels for classification of texture and object categories: a comprehensive study," *International Journal of Computer Vision*, vol. 73, no. 2, pp. 213-238, 2007.

[6] F. Moosmann, B. Triggs, and F. Jurie, "Fast discriminative visual codebooks using randomized clustering forests," in *Neural Information Processing Systems*, 2006, pp. 985-992.

[7] Q. Tian, N. Sebe, M.S. Lew, E. Loupias, and T.S. Huang, "Image Retrieval Using Wavelet-based Salient Points," *J. Electronic Imaging*, vol. 10, no. 4, 2001.

[8] P. Geurts, D. Ernst, and L. Wehenkel, "Extremely randomized trees," *Machine Learning*, vol. 63, no. 1, pp.3-42, 2006.

[9] M. Swain, and D. Ballard, "Color indexing," *International Journal of Computer Vision*, vol. 7, pp. 11-32, 1991.

[10] M. A. Fischler, and R. C. Bolles, "Random sample consensus: a paradigm for model fitting with applications to image analysis and automated cartography," *Communications of the ACM*, vol. 24, no. 6, pp. 381-395, 1981.

[11] J. S. Beis, and D. G. Lowe, "Shape indexing using approximate nearest-neighbor search in high-dimensional spaces," in *Conference on Computer Vision and Pattern Recognition,* pp. 1000-1006, 1997 .

[12] N. Sebe, M.S. Lew, Y. Sun, I. Cohen, T. Gevers, and T.S. Huang, "Authentic Facial Expression Analysis," *Image and Vision Computing*, vol. 25, no. 12, 2007.

[13] N. Sebe and M.S. Lew, "Toward Improved Ranking Metrics," IEEE Transactions on Pattern Analysis and Machine Intelligence, pp. 1132-1143, 2000.